\documentclass[10pt,twocolumn,letterpaper]{article}

\usepackage[pagenumbers]{cvpr} 
\usepackage{colortbl}
\usepackage{color,xcolor}
\usepackage{booktabs}
\usepackage{graphicx}
\usepackage{multirow}
\usepackage{adjustbox}
\usepackage{booktabs}

\definecolor{deepblue}{rgb}{0.7, 0.85, 1}
\definecolor{lightblue}{rgb}{0.6, 0.8, 1}  

\definecolor{cvprblue}{rgb}{0.21,0.49,0.74}
\usepackage[pagebackref,breaklinks,colorlinks,allcolors=cvprblue]{hyperref}

\def\paperID{14446} 
\def\confName{CVPR}
\def\confYear{2025}

\title{Multi-Layer Visual Feature Fusion in Multimodal LLMs: \\Methods, Analysis, and Best Practices}
\author{%
Junyan Lin$^{1,2*}$ \quad Haoran Chen$^{1,3*}$ \quad Yue Fan$^{4}$ \quad Yingqi Fan$^1$ \\
{Xin Jin}$^{1,\dagger}$ \quad {Hui Su}$^5$ \quad {Jinlan Fu}$^{7,\dagger}$ \quad {Xiaoyu Shen}$^{1,6}$\\
$^1$Ningbo Key Laboratory of Spatial Intelligence and Digital Derivative, Institute of Digital Twin, EIT  \\
$^2$Ocean University of China \quad $^3$Zhejiang Gongshang University \quad $^4$Genmo.ai \quad $^5$Meituan Inc.  \\
$^6$Engineering Research Center of Chiplet Design and Manufacturing of Zhejiang Province \quad $^7$NUS  \\
}

\begin{document}
\newcommand\blfootnote[1]{%
\begingroup
\renewcommand\thefootnote{}\footnote{#1}%
\addtocounter{footnote}{-1}%
\endgroup
}

\maketitle
\begin{abstract}

Multimodal Large Language Models (MLLMs) have made significant advancements in recent years, with visual features playing an increasingly critical role in enhancing model performance. However, the integration of multi-layer visual features in MLLMs remains underexplored, particularly with regard to optimal layer selection and fusion strategies. Existing methods often rely on arbitrary design choices, leading to suboptimal outcomes. In this paper, we systematically investigate two core aspects of multi-layer visual feature fusion: (1) selecting the most effective visual layers and (2) identifying the best fusion approach with the language model. Our experiments reveal that while combining visual features from multiple stages improves generalization, incorporating additional features from the same stage typically leads to diminished performance. Furthermore, we find that direct fusion of multi-layer visual features at the input stage consistently yields superior and more stable performance across various configurations. We make all our code publicly available: \url{https://github.com/EIT-NLP/Layer_Select_Fuse_for_MLLM}.

\end{abstract}

\section{Introduction}
\label{sec:intro}
\blfootnote{$*$ Equal contribution. $^\dagger$ Corresponding authors.}

\begin{figure}[t]
  \raggedright 
  \includegraphics[width=\linewidth]{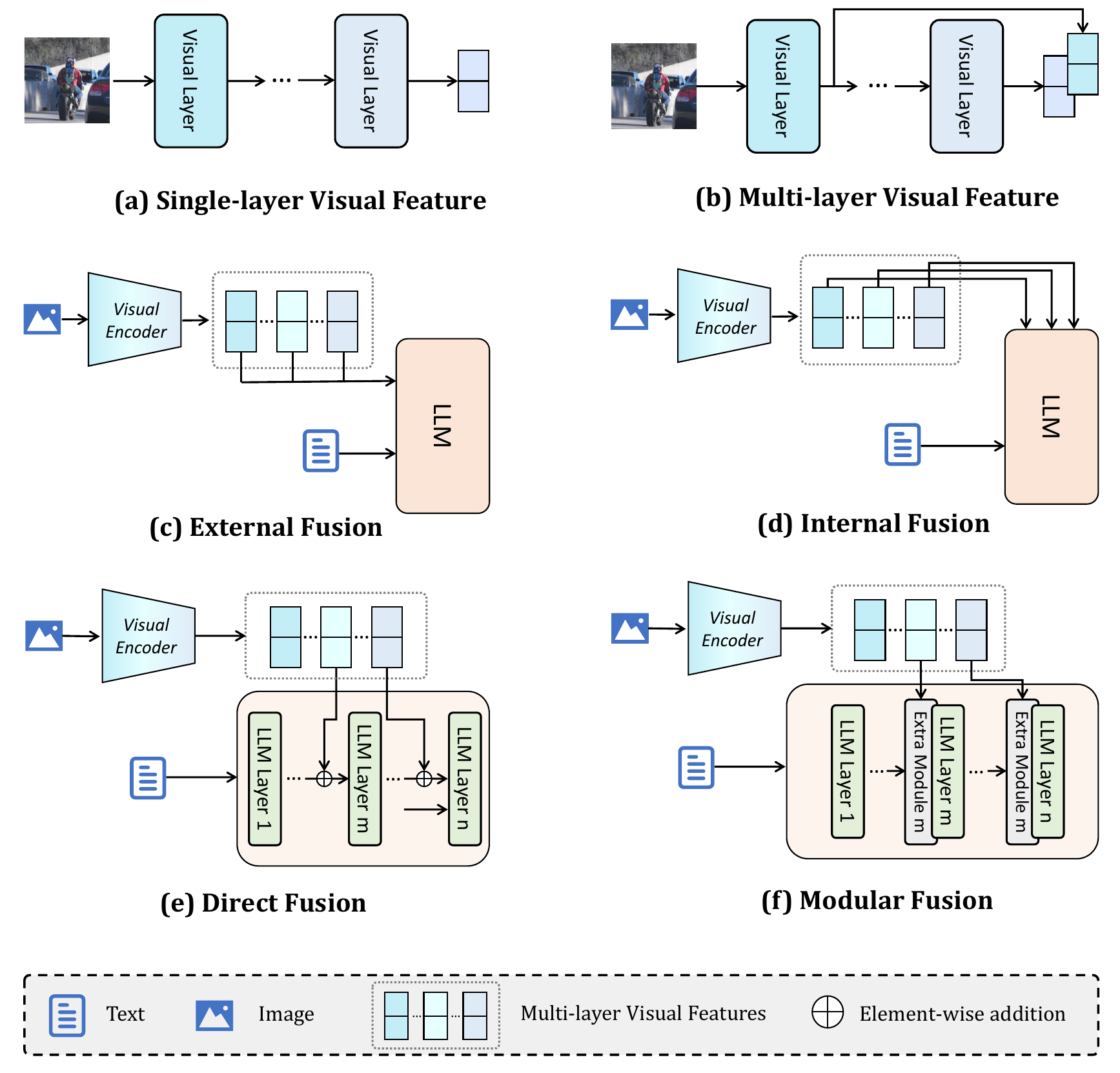}
  \caption{Different Visual Features and Fusion Paradigms. (a) and (b) illustrate the acquisition methods for single-layer and multi-layer visual features, respectively. (c), (d), (e), and (f) display four different fusion strategies: the first two categorize fusion strategies based on fusion position, while the latter two classify fusion strategies based on fusion pattern.}

  \label{fig:fig1}
  \vspace{-10pt}
\end{figure}

Multimodal Large Language Models (MLLMs)~\cite{liu2024llava} have recently achieved impressive results across a range of multimodal tasks, such as image captioning~\cite{ke2019reflective} and visual question answering (VQA)~\cite{antol2015vqa}, by combining pretrained visual encoders~\cite{radford2021learning} with Large Language Models (LLMs)~\cite{dubey2024llama}. 
While substantial research has focused on LLMs in this framework, the visual components, despite their importance, remain relatively understudied. 
In particular, current approaches lack systematic methods for selecting the optimal visual
layers and integrating visual features into LLMs.

Regarding optimal visual selection, current works debate the use of single-layer (\cref{fig:fig1}-(a)) versus multi-layer visual features (\cref{fig:fig1}-(b)).
On the one hand, models such as MiniCPM \cite{hu2024minicpm}, LLaVA \cite{liu2024visual}, and InternVL \cite{chen2024internvl} achieve strong performance by relying on single-layer visual features. 
On the other hand, empirical studies \cite{yao2024dense,chen2024evlm,chen2024lion} have shown that multi-layer visual features can enhance model performance. However, they often select multi-layer visual features in an arbitrary manner. For instance, Dense Connector \cite{yao2024dense} selects layers proportionally based on the depth of the visual encoder, and EVLM \cite{chen2024evlm} directly utilizes features from the latter half of the layers. 
Despite the promising results, there is no systematic method for selecting the optimal visual layers.
This raises the research question: \textbf{How can we select visual layers more effectively?}

To address this question, we propose two criteria for dividing visual layers: Similarity-based and proportion-based selection. Similarity-based selection is inspired by \cite{sun2024transformer}, where visual layers are divided into three groups (beginning, middle, and ending) based on the cosine similarity of visual features from different layers, with each group sharing similar representations and information. Proportion-based selection, on the other hand, is based on the simple proportional selection of visual features, which aligns with many current works using multi-layer visual features. We divide the visual features into three groups (former, latter, and all). Guided by these two criteria, we conducted extensive experiments to explore the optimal layer selection set. Our experiments show that selecting a single representative visual feature from the beginning, middle, and ending stages yields the strongest generalization performance.

Regarding the fusion strategy—how visual features are integrated with LLMs, existing methods also vary widely: some integrate visual features into the intermediate layers of LLMs \cite{hong2024cogagent,meng2024deepstack,alayrac2022flamingo,ye2024mplug}, while others input them at the beginning alongside text features \cite{liu2024visual,liu2024llava,zhu2023minigpt,yuan2024osprey}. Furthermore, approaches differ in whether they use additional modules to process visual features before fusion \cite{chen2024evlm,cao2024mmfuser,hong2024cogagent,chen2024lion} or directly incorporate visual information into the LLM without extra components \cite{liu2024improved,meng2024deepstack,yao2024dense,lin2023sphinx}. The lack of systematic study into fusion strategies leaves a significant gap in understanding how different fusion patterns and positions impact model performance. Consequently, another important question is: 
\textbf{How can we select effective fusion strategies?}

To address these gaps, we first categorize fusion approaches according to two criteria: fusion position and fusion pattern. Fusion position refers to where in the model visual features are integrated—either at the input (external fusion in \cref{fig:fig1}-(c)) or within intermediate layers (internal fusion in \cref{fig:fig1}-(d)). Fusion pattern, on the other hand, distinguishes between modular fusion (\cref{fig:fig1}-(f)), which introduces additional modules for processing visual information, and direct fusion (\cref{fig:fig1}-(e)), which does not require extra components. Guided by this categorization, we develop four fusion strategies and conduct extensive experiments to identify the most effective approach for integrating visual information in MLLMs. Our results reveal that external direct fusion consistently performs best, offering stable and superior results across various configurations. Additionally, we highlight the potential of internal direct fusion, particularly for models trained on large datasets. 

In summary, we provide a comprehensive investigation into visual layer selection and fusion strategy design, offering valuable insights into the effective utilization of multi-layer visual features in MLLMs. Through extensive experiments, we uncover several key findings. Specifically, for visual layer selection, we examine two criteria—representational similarity and layer ratio—and discover that the best performance is achieved when visual features are drawn from distinct representational similarity stages. Moreover, using multiple features from the same stage leads to a performance decline. For fusion strategy selection, we develop four fusion approaches based on fusion position and pattern, and demonstrate that external direct fusion delivers the strongest generalization performance across a variety of configurations.

\section{Related Work}
\label{sec:related}

\subsection{Large Language Model}
Large Language Models (LLMs) are typically trained using an autoregressive method, where the model predicts the next token in a sequence. These models can have billions or even hundreds of billions of parameters and are trained on datasets containing trillions of tokens. For example, LLaMA 3.1 \cite{dubey2024llama} was trained on over 15 trillion tokens. Current LLMs \cite{achiam2023gpt,zhang2022opt,bai2023qwen,jiang2023mistral,team2023internlm} demonstrate exceptional performance, and find widespread application in various fields. However, many of these applications have an increasing demand for efficient large models capable of running on edge devices, driving the need for smaller, more efficient models. Several works \cite{abdin2024phi,chu2023mobilevlm,team2024gemma,hu2024minicpm} have shown that models with parameter sizes as small as 7 billion can still achieve strong performance, meeting the growing need for compact and efficient solutions.

\subsection{Multimodal Large Language Model}

The impressive performance of LLMs has spurred the exploration of MLLMs, leading to the emergence of many outstanding works \cite{liu2024visual,zhu2023minigpt,li2023blip} in recent years. Although these models are highly effective, the information provided by image features from the last layer may be limited. As a result, many methods incorporate additional information in an attempt to better leverage the reasoning capabilities of LLMs. Some works \cite{li2024monkey,zhang2024internlm,meng2024deepstack,hu2024mplug} incorporate multiple high-resolution image patches in addition to a low-resolution image. DeepStack \cite{meng2024deepstack} partitions high-resolution images into a fixed number of sub-images and adds their tokens to the visual tokens between LLM layers.

Some works \cite{rasheed2024glamm,lai2024lisa,yuan2024osprey,you2023ferret} incorporate additional object-level visual information based on the global image. GLAMM \cite{rasheed2024glamm}, based on global features extracted from a low-resolution image, adds a region encoder to achieve local feature extraction. 
Some works \cite{lin2023sphinx,jiang2023clip,zhang2024ferret} utilize visual encoders pretrained with different methods to extract various visual features, combining these features as visual tokens to the LLM. SPHINX \cite{lin2023sphinx} uses multiple visual encoders to extract different visual information, enriching the details. 
Some works \cite{yao2024dense,hong2024cogagent,chen2024evlm,chen2024lion} leverage multi-layer visual features. EVLM \cite{chen2024evlm} utilizes hierarchical ViT features for enabling the model to perceive visual signals as comprehensively as possible.

Although there have been various attempts to integrate extra information into LLM, there is a lack of extensive exploration on how to effectively fuse this information. Moreover, many of these approaches rely on expanded datasets or more intricate architectures compared to baselines, making it unclear whether observed improvements result from advanced fusion strategies or simply from increased model capacity. To address these issues, we systematically investigate the optimal fusion paradigm by leveraging multi-layer visual features.

\section{Methodology}

While many works \cite{chen2024evlm,yao2024dense,cao2024mmfuser,chen2024lion} demonstrate that incorporating multi-layer visual features can significantly improve MLLM performance, \textit{they do not explore the deeper rationale behind effective layer selection, nor do they generalize across diverse fusion strategies.}
Next, we provide a detailed illustration of visual layer selection (Sec.~\ref{sec:layer-selction}) and the two fusion strategies, namely internal fusion (Sec.~\ref{sec:internal}) and external fusion (Sec.~\ref{sec:external}), to address Research Questions 1 and 2, as demonstrated in Sec.~\ref{sec:intro}.

\subsection{Visual Layer Selection}
\label{sec:layer-selction}
\begin{figure}[htpb]
    \centering
  \includegraphics[width=\columnwidth]{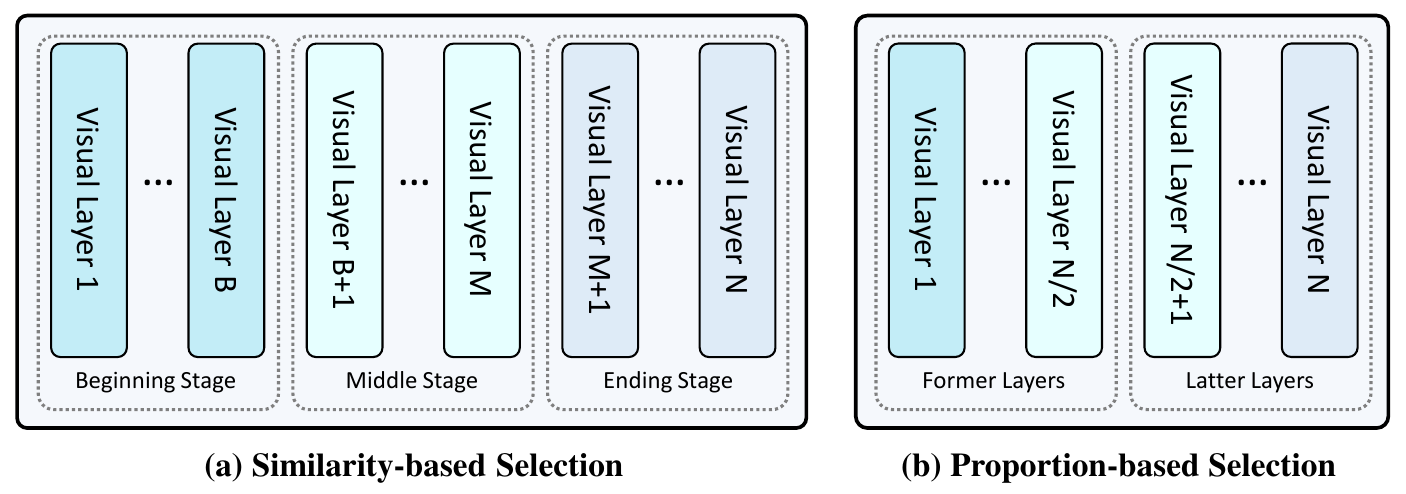}
  \caption{
Comparison of Similarity-Based and Proportion-Based Visual Layer Selection.
  }
  \label{fig:layer_selection}
\end{figure}

We categorize visual layers using two criteria: \textbf{similarity-based selection} and \textbf{proportion-based selection}. The similarity-based selection approach is motivated by findings from studies \cite{sun2024transformer, raghu2021vision}, which show that features within the same stage (beginning, middle, or ending) often reside in a similar representation space and share comparable properties. This similarity allows us to divide the layers into meaningful stages, selecting representative layers from each to capture diverse types of visual information. As illustrated in \cref{fig:layer_selection}-(a), the similarity-based selection approach divides the $N$ layers of a visual encoder into three groups: layers $1$ through $B$ represent the beginning stage, layers $B+1$ through $M$ represent the middle stage, and layers $M+1$ through $N$ represent the ending stage.

\begin{figure*}[!t]
    \centering
  \includegraphics[width=\textwidth]{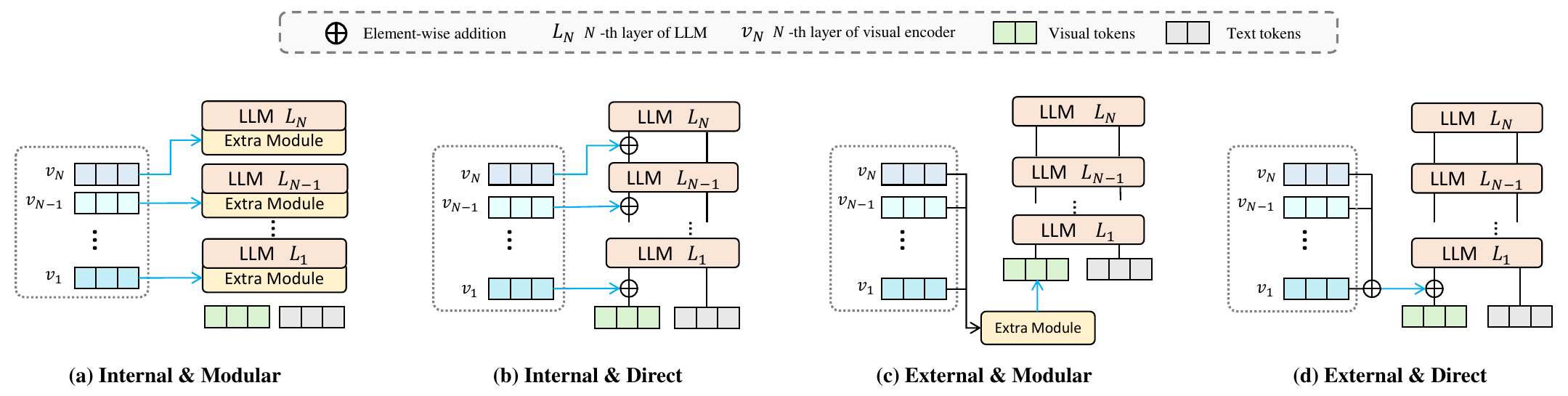}
  \caption{
  Framework of the four fusion strategies studied in this work.
 \textcolor[RGB]{0,176,240}{Blue} lines represent the path passing through the projector. 
  }
  \label{fig:different_fusion}
\end{figure*}

On the other hand, the proportion-based selection strategy aligns with the previous works \cite{chen2024evlm,chen2024lion,yao2024dense,cao2024mmfuser}, where layers are selected based on a proportional division of the encoder's depth. To ensure consistency with prior research and facilitate a systematic comparison, we apply this method by dividing the $N$ layers of the visual encoder into two groups: the \textit{former} and \textit{latter} stages. As shown in \cref{fig:layer_selection}-(b), layers $1$ through $N/2$ constitute the former stage, while layers $N/2+1$ through $N$ form the latter stage. This approach thus separately examines the contributions of shallow and deep features across the encoder's depth.

For similarity-based selection, we choose representative layers based on empirical findings\footnote{For each stage, we traine each layer based on LLaVA setting (substituting the Vicuna 1.5 7B with the MobileLLaMA 1.4B) and selected the layer with the highest average performance as the representative.}: the 3rd layer (beginning stage), 18th layer (middle stage), and 23rd layer (ending stage), resulting in three sets: \{18\}, \{3, 18\}, and \{3, 18, 23\}. For proportion-based selection, we define three sets as well: \{former\}, \{latter\}, and \{all\}. These sets allow for a systematic evaluation of how layer selection influences model performance across different aspects.

\subsection{Internal Fusion}
\label{sec:internal}

Internal fusion methods incorporate tokens containing additional information directly within the LLM. 
Given a visual feature set (multi-layer visual features) $\mathbf{F} = \{v_1, v_2, \ldots, v_N\} \footnote{In internal fusion, for simplicity, we assume that the number of selected visual layers and LLM layers are both $N$}\in \mathbb{R}^{N_v \times d}$, where $N_v$ represents number of visual patches and $d$ represents the channel dimension, and hidden states $\mathbf{H}= \{h_1, h_2, \ldots, h_N\} \in \mathbb{R}^{N_l \times D}$ in the LLM, where $N_l$ denotes the number of tokens and $D$ is the hidden size of the LLM, the operation for the internal fusion $\mathcal{I}$ at layer $i$ within the LLM can be formulated as:
\begin{equation}
h_i^{'} = \mathcal{I}(h_i, \mathbf{P}_i(v_i)) + h_i, 
\end{equation}
where $h_i^{'} \in \mathbb{R}^{N_l \times D}$ denotes the updated hidden state in layer $i$, and $\mathbf{P}_i$ is a projector at layer $i$ that aligns the visual feature $v_i$ with the LLM's embedding space.

As shown in \cref{fig:different_fusion}-(a) and \cref{fig:different_fusion}-(b), the distinction between \textbf{Internal Modular fusion} and \textbf{Internal Direct fusion} lies in the differences in using $\mathcal{I}$. In internal modular fusion, cross-attention modules are the most commonly used modules for integrating multi-layer visual features. Depending on where the cross-attention is inserted, this method can be further divided into pre-cross attention, post-cross attention, and parallel attention architectures \cite{ye2024mplug}.
Internal Direct fusion, on the other hand, simply integrates multi-layer visual information by directly adding the visual tokens at their respective positions.

\subsection{External Fusion}
\label{sec:external}

External fusion methods integrate multi-layer visual features at the input stage before feeding visual tokens into the LLMs. 
Given a visual feature set $\mathbf{F}$, visual tokens $\mathbf{V}\in \mathbb{R}^{N_v \times D}$, and text tokens $\mathbf{T}\in \mathbb{R}^{N_t \times D}$, where $N_t$ represents the number of text tokens, the operation for external fusion $\mathcal{E}$ can be formulated as:
\begin{equation}
\mathbf{V}' = \mathcal{E}(\mathbf{V},\mathbf{P}(\mathbf{F})), 
\end{equation}
where $\mathbf{V}' \in \mathbb{R}^{N_v \times D}$ denotes the updated visual tokens after the external fusion operation. As shown in \cref{fig:different_fusion}-(c) and \cref{fig:different_fusion}-(d), similar to internal fusion, the distinction between \textbf{External Modular Fusion} and \textbf{External Direct Fusion} lies in how $\mathcal{E}$ is applied. In external modular fusion, using various modules, multi-layer visual features $\mathbf{F}$ are integrated into $\mathbf{V}$ to get updated visual tokens $\mathbf{V}'$, or directly generate $\mathbf{V}'$ by multi-layer visual features $\mathbf{F}$. External direct fusion, on the other hand, combines $\mathbf{F}$ with $\mathbf{V}$ through simpler operations like element-wise addition, stacking along the $N$ dimension, or stacking along the $D$ dimension.

Notably, in internal fusion, each visual feature requires a new projector when introduced at different LLM layers, resulting in increased parameters as the number of layers grows. In contrast, external fusion only requires a single projector for every visual layer sets, making it more parameter-efficient when dealing with multiple layers.

This paper will provide a detailed analysis of existing fusion methods, explore their adaptability of handling extra multi-layer visual features, and experimentally validate the strengths and weaknesses of each method, offering valuable insights for future research and applications.

\subsection{Base Model: Mini-LLaVA}
To facilitate the deployment of our exploratory experiments, we developed a lightweight MLLM, namely Mini-LLaVA, based on modifications to LLaVA-1.5.

\noindent
\textbf{Structure} Mini-LLaVA replaces the Vicuna 1.5 (7B) model \cite{vicuna} with a smaller mobileLLaMA (1.4B) model \cite{chu2023mobilevlm} while keeping the other components consistent with LLaVA-1.5. The main exploration experiments in this paper are conducted based on Mini-LLaVA.
More specifically, like LLaVA-1.5, Mini-LLaVA uses the feature from the 23rd layer of the visual encoder, which is fed into the LLM. It employs the 24-layer MobileLLaMA 1.4B as the LLM and CLIP-ViT-L/14 \cite{radford2021learning} as the visual encoder. Since both the visual encoder and the LLM have 24 layers, we adopt a layer-wise alignment method for internal fusion, where each visual feature layer can fuse with the corresponding layer in the LLM (e.g., the visual features of the 18th layer are fused into the 18th layer of the LLM).

\vspace{6pt}
\noindent
\textbf{Training} Consistent with LLaVA-1.5, we use a pre-training stage with a dataset comprising 558K image captions \cite{schuhmann2022laion,changpinyo2021conceptual,yun2012two}, and an instruction tuning stage with a dataset of 665K conversations \cite{liu2024visual,liu2024improved}. During the pre-training stage, only newly initialized components, such as the projector and any new modules within modular fusion, are trained. In the instruction tuning stage, all parameters except the visual encoder are unfrozen and optimized.

\section{Experiment and Analysis}
\label{sec:experiment}

\subsection{Experiment Settings}

\noindent
\textbf{Comparing Models}
We design Mini-LLaVA, which utilizes visual features from layer 23. We then compare the performance of this model when fusing features from different individual visual layers or sets of layers. Specifically, the visual layer sets considered include \textit{Single}:\{18\}, \textit{Double}:\{3, 18\}, \textit{Triple}:\{3, 18, 23\}, \textit{Former}: \{former\}, \textit{Latter}: \{latter\}, and \textit{All}: \{all\}, as introduced in Sec.~\ref{sec:layer-selction}.

\vspace{6pt}
\noindent
\textbf{Benchmarks}
To conduct a comprehensive evaluation of performance, we evaluate different settings across four benchmark categories: \textit{General}, \textit{OCR}, \textit{CV-Centric}, and \textit{Hallucination (Hallu)}. The \textit{General} category includes GQA \cite{hudson2019gqa}, MMBench (MMB) \cite{liu2025mmbench}, and MME \cite{fu2023mme}, which is further divided into MME Cognition (MME$^C$) and MME Perception (MME$^P$). The \textit{OCR} category covers TextVQA \cite{singh2019textvqa} and OCRBench \cite{liu2023ocrbench}. In the \textit{CV-Centric} category, we include CV-Bench \cite{tong2024cambrian}, which itself contains two subcategories: CV-Bench 2D and CV-Bench 3D. Finally, the \textit{Hallucination} category is represented by POPE \cite{li2023pope}.

\subsection{Internal Fusion Ablation}

Modular fusion in internal fusion can take several forms: pre-cross, post-cross, and parallel, which involve a higher level of complexity compared to direct fusion. Therefore, we start by investigating these configurations and then discuss direct fusion to explore the differences between them.

\subsubsection{Modular Fusion}

\begin{table*}[!t]
    \centering
    \caption{Results on Pre-Cross Attention Fusion. \textbf{Note:} MME$^{C}$ represents MME Cognition, and MME$^{P}$ represents MME Perception. The superscript numbers in the top right indicate the score difference compared to the baseline, Mini-LLaVA. \textit{Single} (\{18\}), \textit{Double} (\{3, 18\}), \textit{Triple} (\{3, 18, 24\}), \textit{Former} ([1, ..., 12]), \textit{Latter} ([13, ..., 24]), and \textit{All}  ([1, ..., 24]) are the visual layer sets that the Mini-LLaVA adopted.}
    \small
    \renewcommand\tabcolsep{2.3pt}
    \begin{tabular}{lllllllllll}
    \toprule
    \multirow{2}[2]{*}{\textbf{Model}} & \multicolumn{4}{c}{\textbf{General}} & \multicolumn{2}{c}{\textbf{OCR}} & \multicolumn{2}{c}{\textbf{CV-Centric}} & \textbf{Hallu} & \multirow{2}[2]{*}{\textbf{Avg.}} \\ 
    \cmidrule(lr){2-5}\cmidrule(r){6-7} \cmidrule(r){8-9} \cmidrule(r){10-10}
    & \textbf{GQA} & \textbf{MMB} & \textbf{MME$^{C}$} & \textbf{MME$^{P}$} & \textbf{TextVQA} & \textbf{OCRBench} & \textbf{CVBench$^{2D}$} & \textbf{CVBench$^{3D}$} & \textbf{POPE} & \\ 
    \midrule
    \rowcolor{gray!15} Mini-LLaVA                  & 56.95 & 46.91 & 262 & 1200 & 35.47 & 239 & 39.74 & 55.00 & 85.83 & 48.51 \\
    + \textit{Single}                       & 57.89$^{\color{red}{0.94}\uparrow}$ & 50.77$^{\color{red}{3.86}\uparrow}$ & 228$^{\color{teal}{34}\downarrow}$ & 1153$^{\color{teal}{47}\downarrow}$ & 35.04$^{\color{teal}{0.43}\downarrow}$ & 253$^{\color{red}{14}\uparrow}$ & 41.66$^{\color{red}{1.92}\uparrow}$ & 54.58$^{\color{teal}{0.42}\downarrow}$ & 86.03$^{\color{red}{0.20}\uparrow}$ & 48.60$^{\color{red}{0.09}\uparrow}$ \\
    + \textit{Double}                     & 58.41$^{\color{red}{1.46}\uparrow}$ & 50.93$^{\color{red}{4.02}\uparrow}$ & 218$^{\color{teal}{44}\downarrow}$ & 1182$^{\color{teal}{18}\downarrow}$ & 34.42$^{\color{teal}{1.05}\downarrow}$ & 261$^{\color{red}{22}\uparrow}$ & 46.34$^{\color{red}{6.60}\uparrow}$ & 51.08$^{\color{teal}{3.92}\downarrow}$ & 85.06$^{\color{teal}{0.77}\downarrow}$ & 48.74$^{\color{red}{0.23}\uparrow}$ \\
    + \textit{Triple}                 & 57.56$^{\color{red}{0.61}\uparrow}$ & 49.66$^{\color{red}{2.75}\uparrow}$ & 212$^{\color{teal}{50}\downarrow}$ & 1163$^{\color{teal}{37}\downarrow}$ & 34.06$^{\color{teal}{1.41}\downarrow}$ & 255$^{\color{red}{16}\uparrow}$ & 38.66$^{\color{teal}{1.08}\downarrow}$ & 47.42$^{\color{teal}{7.58}\downarrow}$ & 84.69$^{\color{teal}{1.14}\downarrow}$ & 46.91$^{\color{teal}{1.60}\downarrow}$ \\
    + \textit{Former}                    & 55.91$^{\color{teal}{1.04}\downarrow}$ & 45.70$^{\color{teal}{1.21}\downarrow}$ & 227$^{\color{teal}{35}\downarrow}$ & 1162$^{\color{teal}{38}\downarrow}$ & 30.71$^{\color{teal}{4.76}\downarrow}$ & 165$^{\color{teal}{74}\downarrow}$ & 38.68$^{\color{teal}{1.06}\downarrow}$ & 51.42$^{\color{teal}{3.58}\downarrow}$ & 85.03$^{\color{teal}{0.80}\downarrow}$ & 45.60$^{\color{teal}{2.91}\downarrow}$ \\
    + \textit{Latter}                  & 49.92$^{\color{teal}{7.03}\downarrow}$ & 0.17$^{\color{teal}{46.74}\downarrow}$ & 250$^{\color{teal}{12}\downarrow}$ & 906$^{\color{teal}{294}\downarrow}$ & 18.96$^{\color{teal}{16.51}\downarrow}$ & 136$^{\color{teal}{103}\downarrow}$ & 44.51$^{\color{red}{4.77}\uparrow}$ & 48.83$^{\color{teal}{6.17}\downarrow}$ & 82.13$^{\color{teal}{3.70}\downarrow}$ & 37.19$^{\color{teal}{11.32}\downarrow}$ \\
    + \textit{All}                    & --    & --     & --     & --    & --  & --   & --  & --    & --    & -- \\
    \bottomrule
    \end{tabular}
    \label{tab: ablation on Cross-self}
\end{table*}

\paragraph{Layer Combination Exploration:} We first conduct an extensive investigation into the pre-cross fusion because of its commonality. The performance of pre-cross attention fusion across six layer selection sets is summarized in \cref{tab: ablation on Cross-self}. Due to convergence difficulties when applying \textit{All}, we are unable to complete an evaluation for this configuration. Several key insights emerge from the results:

\begin{figure}[!t]
    \centering
  \includegraphics[width=0.8\columnwidth]{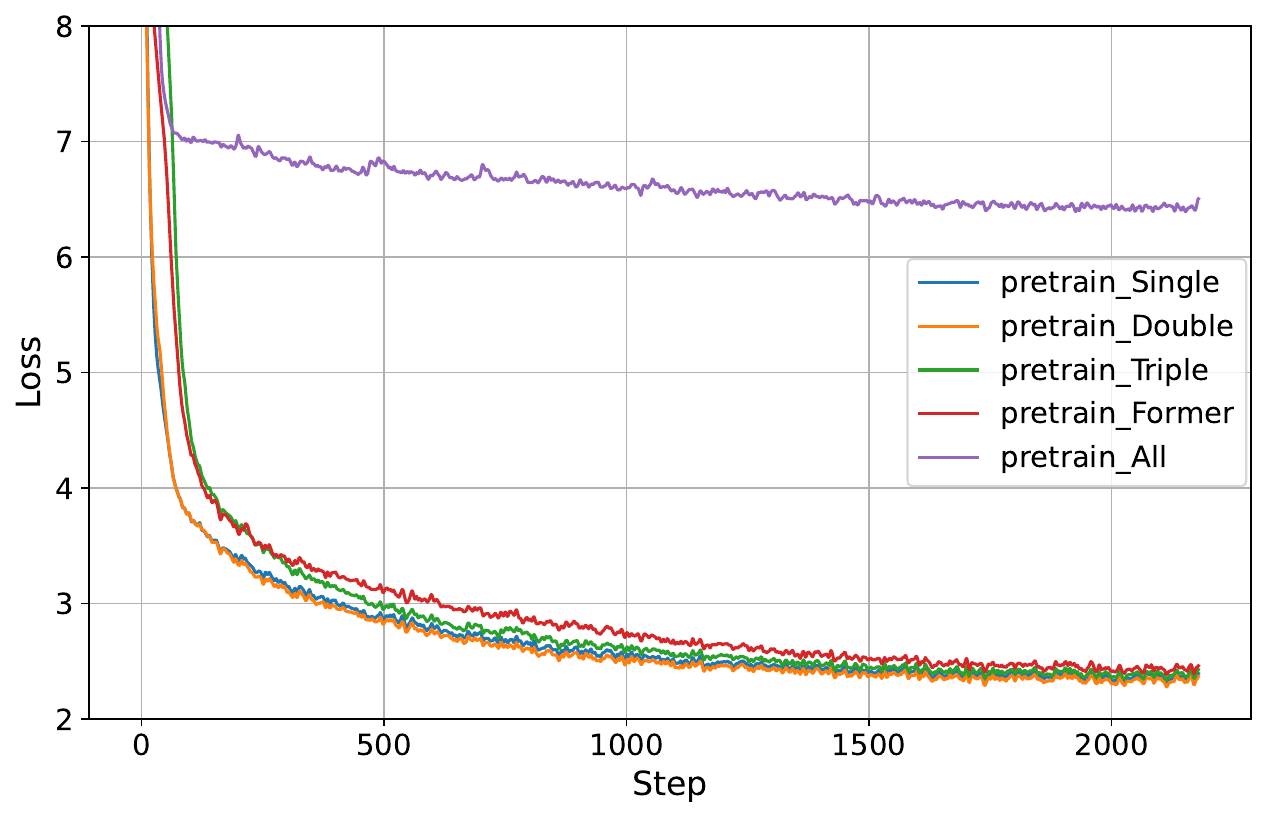}
  \vspace{-6pt}
  \caption{
Pre-cross attention loss curves in pre-training stage under different layer sets.
  }
  \label{fig:pretrain_loss}
\end{figure}

\begin{itemize}
    \item \textbf{Beginning-Stage Features Benefit Detail-Specific Tasks}: The \textit{Double} shows improved performance over \textit{Single} alone in tasks requiring image details. This highlights the importance of beginning-stage features in enhancing detail-specific aspects, making it advantageous to incorporate both beginning and middle-stage layers for comprehensive feature representation.

    \item \textbf{Limitations of Extensive Layer Selection}: Inserting an excessive number of modules can lead to significant performance degradation, primarily due to the difficulty in optimizing numerous parameters. As shown in \cref{fig:pretrain_loss}, which illustrates the loss curves of pre-training under different layer sets, configurations with more layers tend to encounter greater training challenges. In particular, the \textit{All} displayed convergence issues, with the training loss plateauing around 6 after an initial drop from 11, suggesting instability in the training process.
    
    \item \textbf{Performance Gap Between \textit{Former} and \textit{Latter}}: Performance declines substantially when additional modules are inserted into latter layers compared to former layers, despite similar parameter counts. This suggests that inserting modules into latter layers may disrupt the model's feature processing. When modules are added early, they allow subsequent layers to refine and correct features, which is less feasible with latter layer insertions. This is further illustrated by the performance gap between \textit{Double} and \textit{Triple}, with the former showing better results.

    \item \textbf{Limited Improvement on Performance}: Among the five sets, only \textit{Double} and \textit{Triple} show slight improvements over the baseline, outperforming it by just 0.09 and 0.23 points, respectively, while other configurations actually perform worse.

\end{itemize}

\begin{table*}[!t]
    \centering
 \caption{Comparison on Different Modular Fusion Architectures. The superscript numbers in the top right indicate the score difference compared to the Pre-Cross Fusion.}
    \label{tab: ablation on modular fusion strategies}
     \small
    \renewcommand\tabcolsep{2.3pt}
    \begin{tabular}{lllllllllll}
    \toprule
    \multirow{2}[2]{*}{\textbf{Architectures}} & \multicolumn{4}{c}{\textbf{General}} & \multicolumn{2}{c}{\textbf{OCR}} & \multicolumn{2}{c}{\textbf{CV-Centric}} & \textbf{Hallu} & \multirow{2}[2]{*}{\textbf{Avg.}} \\ 
    \cmidrule(lr){2-5}\cmidrule(r){6-7}\cmidrule(r){8-9}\cmidrule(r){10-10} 
    & \textbf{GQA} & \textbf{MMB} & \textbf{MME$^{C}$} & \textbf{MME$^{P}$} & \textbf{TextVQA} & \textbf{OCRBench} & \textbf{CVBench$^{2D}$} & \textbf{CVBench$^{3D}$} & \textbf{POPE} & \\ 
    \midrule
    Pre-Cross                        & 58.41 & 50.93 & 218 & 1182 & 34.42 & 261 & 46.34 & 51.08 & 85.06 & 48.74 \\
    Post-Cross                      & 57.87$^{\color{teal}{0.54}\downarrow}$ & 50.43$^{\color{teal}{0.50}\downarrow}$ & 254$^{\color{red}{36}\uparrow}$ & 1171$^{\color{teal}{11}\downarrow}$ & 34.68$^{\color{red}{0.26}\uparrow}$ & 244$^{\color{teal}{17}\downarrow}$ & 44.80$^{\color{teal}{1.54}\downarrow}$ & 54.92$^{\color{red}{3.84}\uparrow}$ & 85.80$^{\color{red}{0.74}\uparrow}$ & 49.24$^{\color{red}{0.50}\uparrow}$ \\
    Parallel                   & 58.05$^{\color{teal}{0.36}\downarrow}$ & 49.74$^{\color{teal}{1.19}\downarrow}$ & 218$^{\color{gray}{0}}$ & 1191$^{\color{red}{9}\uparrow}$ & 34.70$^{\color{red}{0.28}\uparrow}$ & 239$^{\color{teal}{22}\downarrow}$ & 44.57$^{\color{teal}{1.77}\downarrow}$ & 52.92$^{\color{red}{1.84}\uparrow}$ & 85.20$^{\color{red}{0.14}\uparrow}$ & 48.43$^{\color{teal}{0.31}\downarrow}$ \\
    \bottomrule
    \end{tabular}
\end{table*}

\paragraph{Different Modular Fusion Strategies:} To explore the differences in performance across pre-cross attention, post-cross attention, and parallel attention architectures,
we utilize the layer selection set \textit{Double}, which showed optimal results for pre-cross attention, and test it under post-cross attention and parallel attention architectures to evaluate their performance. The results are summarized in \cref{tab: ablation on modular fusion strategies}. Although the three fusion strategies each show specific advantages on different benchmarks, their overall performance remains close.
This suggests that, for multi-layer visual feature fusion, the choice of modular fusion strategy has relatively low significance.


\subsubsection{Direct Fusion}

\begin{table*}[!h]
    \centering
     \caption{Results of Internal Direct Fusion. The superscript numbers in the top right indicate the metric difference compared to the Internal Modular Fusion.}
    \label{tab: ablation on fusion strategies compare}
   \small
    \renewcommand\tabcolsep{2.1pt}
    \begin{tabular}{lllllllllll}
    \toprule
    \multirow{2}[2]{*}{\textbf{Model}} & \multicolumn{4}{c}{\textbf{General}} & \multicolumn{2}{c}{\textbf{OCR}} & \multicolumn{2}{c}{\textbf{CV-Centric}} & \textbf{Hallu} & \multirow{2}[2]{*}{\textbf{Avg.}} \\ 
    \cmidrule(lr){2-5}\cmidrule(r){6-7}\cmidrule(r){8-9}\cmidrule(r){10-10}
    & \textbf{GQA} & \textbf{MMB} & \textbf{MME$^{C}$} & \textbf{MME$^{P}$} & \textbf{TextVQA} & \textbf{OCRBench} & \textbf{CVBench$^{2D}$} & \textbf{CVBench$^{3D}$} & \textbf{POPE} & \\ 
    \midrule
    \rowcolor{gray!15} Mini-LLaVA & 56.95 & 46.91 & \textbf{262} & 1200 & 35.47 & 239 & 39.74 & 55.00 & 85.83 & 48.51 \\

    + \textit{Single} & 58.08$^{\color{red}{0.19}\uparrow}$ & 52.41$^{\color{red}{1.64}\uparrow}$ & 234$^{\color{red}{6}\uparrow}$ & 1154$^{\color{red}{1}\uparrow}$ & 36.03$^{\color{red}{0.99}\uparrow}$ & 251$^{\color{teal}{2}\downarrow}$ & 41.15$^{\color{teal}{0.51}\downarrow}$ & 52.58$^{\color{teal}{2.00}\downarrow}$ & 85.66$^{\color{teal}{0.37}\downarrow}$ & 48.66$^{\color{red}{0.06}\uparrow}$ \\ 
    + \textit{Double} & 58.08$^{\color{teal}{0.33}\downarrow}$ & 48.56$^{\color{teal}{2.37}\downarrow}$ & 229$^{\color{red}{11}\uparrow}$ & 1178$^{\color{teal}{4}\downarrow}$ & 34.95$^{\color{red}{0.53}\uparrow}$ & 237$^{\color{teal}{24}\downarrow}$ & 40.62$^{\color{teal}{5.72}\downarrow}$ & 56.50$^{\color{red}{5.42}\uparrow}$ & 85.77$^{\color{red}{0.71}\uparrow}$ & 48.41$^{\color{teal}{0.33}\downarrow}$ \\ 
    + \textit{Triple} & 58.59$^{\color{red}{1.03}\uparrow}$ & 47.47$^{\color{teal}{2.19}\downarrow}$ & 223$^{\color{red}{11}\uparrow}$ & 1207$^{\color{red}{44}\uparrow}$ & 36.24$^{\color{red}{2.18}\uparrow}$ & 255$^{\color{gray}0}$ & 41.87$^{\color{red}{3.21}\uparrow}$ & 53.08$^{\color{red}{5.66}\uparrow}$ & 85.87$^{\color{red}{1.18}\uparrow}$ & 48.54$^{\color{red}{1.63}\uparrow}$ \\ 
    + \textit{Former} & 54.08$^{\color{teal}{1.83}\downarrow}$ & 43.14$^{\color{teal}{2.56}\downarrow}$ & 238$^{\color{red}{11}\uparrow}$ & 1120$^{\color{teal}{42}\downarrow}$ & 26.51$^{\color{teal}{4.20}\downarrow}$ & 200$^{\color{red}{35}\uparrow}$ & 34.94$^{\color{teal}{3.74}\downarrow}$ & 51.33$^{\color{teal}{0.09}\downarrow}$ & 84.14$^{\color{teal}{0.89}\downarrow}$ & 44.43$^{\color{teal}{1.17}\downarrow}$ \\ 
    + \textit{Latter} & 58.79$^{\color{red}{8.87}\uparrow}$ & 47.02$^{\color{red}{46.85}\uparrow}$ & 221$^{\color{teal}{29}\downarrow}$ & 1179$^{\color{red}{273}\uparrow}$ & 37.28$^{\color{red}{18.32}\uparrow}$ & 241$^{\color{red}{105}\uparrow}$ & 42.64$^{\color{teal}{1.87}\downarrow}$ & 51.92$^{\color{red}{3.09}\uparrow}$ & 85.57$^{\color{red}{3.44}\uparrow}$ & 48.21$^{\color{red}{2.61}\uparrow}$ \\ 
    + \textit{All} & \textcolor{gray}{58.04} & \textcolor{gray}{47.28} & \textcolor{gray}{224} & \textcolor{gray}{1215} & \textcolor{gray}{34.66} & \textcolor{gray}{243} & \textcolor{gray}{42.86} & \textcolor{gray}{51.46} & \textcolor{gray}{85.19} & \textcolor{gray}{48.06} \\

    \bottomrule
    \end{tabular}
\end{table*}

In the comparison of modular fusion, we conduct an in-depth investigation into the choices of layers and modules, yielding several valuable conclusions. For instance, when an excessive number of layers is introduced, a significant performance drop is observed. However, this drop may be due to the additional parameters brought in by the numerous modules. To minimizing the influence of extra parameters, we conduct further experiments centered on direct fusion. Specifically, we employ the fusion strategy in DeepStack \cite{lu2024deepseek} to evaluate the performance differences across six layer selection sets compared to modular fusion. The detailed results, shown in \cref{tab: ablation on fusion strategies compare}, reveal several key properties:

\begin{itemize}
    \item \textbf{Stable Performance on Increased Visual Layer}: As the number of visual layers increases, the model exhibits stable performance; in fact, it shows improvement on some benchmarks. This suggests that direct fusion can effectively adapt to additional visual information without requiring new modules or extensive training data.

    \item \textbf{Resilience to \textit{Latter}}: Unlike modular fusion, which tends to perform better at \textit{Former}, direct fusion demonstrates superior results at \textit{Latter}, especially in GQA and TextVQA. This discrepancy may be due to the different weights of attention LLMs allocate to visual tokens in different layers \cite{chen2024image}. In \textit{Former}, attention to visual tokens is significantly stronger, potentially causing direct fusing multi-layer visual features into hidden states to create greater disruptions. In contrast, \textit{Latter}, which receive less attention, can incorporate multi-layer visual features more smoothly. This is the opposite of the logic in modular fusion, where adding new modules in \textit{Former} tends to be advantageous. The fact that \textit{Double} outperforms \textit{Single} in modular fusion but not in direct fusion further supports this hypothesis.

\end{itemize}

\subsection{External Fusion Ablation}

\begin{table*}[!h]
    \centering
    \caption{Results of External Modular Fusion. The superscript numbers in the top right indicate the metric difference compared to the Internal Modular Fusion.}
    \label{tab: external_modular}
    \scalebox{0.88}{
    \renewcommand\tabcolsep{2.4pt}
    \begin{tabular}{lllllllllll}
    \toprule
    \multirow{2}[2]{*}{\textbf{Model}} & \multicolumn{4}{c}{\textbf{General}} & \multicolumn{2}{c}{\textbf{OCR}} & \multicolumn{2}{c}{\textbf{CV-Centric}} & \textbf{Hallu} & \multirow{2}[2]{*}{\textbf{Avg.}} \\ 
    \cmidrule(lr){2-5}\cmidrule(r){6-7}\cmidrule(r){8-9}\cmidrule(r){10-10}
    & \textbf{GQA} & \textbf{MMB} & \textbf{MME$^{C}$} & \textbf{MME$^{P}$} & \textbf{TextVQA} & \textbf{OCRBench} & \textbf{CVBench$^{2D}$} & \textbf{CVBench$^{3D}$} & \textbf{POPE} & \\ 
    \midrule
    \rowcolor{gray!15} Mini-LLaVA & 56.95 & 46.91 & 262 & 1200 & 35.47 & 239 & 39.74 & 55.00 & 85.83 & 48.51 \\

    + \textit{Single}                         & 58.55$^{\color{red}{0.66}\uparrow}$ & 53.09$^{\color{red}{2.32}\uparrow}$ & 228$^{\color{gray}{0}}$ & 1172$^{\color{red}{19}\uparrow}$ & 35.42$^{\color{teal}{0.38}\downarrow}$ & 238$^{\color{teal}{15}\downarrow}$ & 42.60$^{\color{red}{0.94}\uparrow}$ & 51.92$^{\color{teal}{2.66}\downarrow}$ & 85.72$^{\color{teal}{0.31}\downarrow}$ & 48.69$^{\color{red}{0.09}\uparrow}$ \\
    + \textit{Double}                     & 58.43$^{\color{red}{0.02}\uparrow}$ & 52.66$^{\color{red}{1.73}\uparrow}$ & 225$^{\color{red}{7}\uparrow}$ & 1173$^{\color{teal}{9}\downarrow}$ & 36.25$^{\color{red}{1.83}\uparrow}$ & 262$^{\color{red}{1}\uparrow}$ & 45.78$^{\color{teal}{0.56}\downarrow}$ & 52.83$^{\color{red}{1.75}\uparrow}$ & 86.03$^{\color{red}{0.97}\uparrow}$ & 49.44$^{\color{red}{0.70}\uparrow}$ \\
    + \textit{Former}                    & 58.37$^{\color{red}{2.46}\uparrow}$ & 54.30$^{\color{red}{8.60}\uparrow}$ & 258$^{\color{red}{31}\uparrow}$ & 1181$^{\color{red}{19}\uparrow}$ & 35.85$^{\color{red}{5.14}\uparrow}$ & 244$^{\color{red}{79}\uparrow}$ & 42.59$^{\color{red}{3.91}\uparrow}$ & 54.92$^{\color{red}{3.50}\uparrow}$ & 86.29$^{\color{red}{1.26}\uparrow}$ & 49.78$^{\color{red}{4.18}\uparrow}$ \\
    + \textit{Latter}                   & 58.51$^{\color{red}{8.59}\uparrow}$ & 51.20$^{\color{red}{51.03}\uparrow}$ & 231$^{\color{teal}{19}\downarrow}$ & 1212$^{\color{red}{306}\uparrow}$ & 36.41$^{\color{red}{17.45}\uparrow}$ & 255$^{\color{red}{119}\uparrow}$ & 40.71$^{\color{teal}{3.80}\downarrow}$ & 52.42$^{\color{red}{3.59}\uparrow}$ & 85.80$^{\color{red}{3.67}\uparrow}$ & 48.89$^{\color{red}{11.70}\uparrow}$ \\
    + \textit{All}                     & \textcolor{gray}{58.57} & \textcolor{gray}{51.46} & \textcolor{gray}{233} & \textcolor{gray}{1211} & \textcolor{gray}{35.28} & \textcolor{gray}{265} & \textcolor{gray}{42.00} & \textcolor{gray}{52.08} & \textcolor{gray}{86.26} & \textcolor{gray}{49.09} \\
        
    \bottomrule
    \end{tabular}
    }
\end{table*}

\begin{table*}[htbp]
    \centering
    \caption{Results of External Direct Fusion. The superscript numbers in the top right indicate the metric difference compared to the Internal Direct Fusion.}
    \label{tab: ablation on external fusion}
    \small
    \renewcommand\tabcolsep{2.4pt}
    \begin{tabular}{lllllllllll}
    \toprule
    \multirow{2}[2]{*}{\textbf{Model}} & \multicolumn{4}{c}{\textbf{General}} & \multicolumn{2}{c}{\textbf{OCR}} & \multicolumn{2}{c}{\textbf{CV-Centric}} & \textbf{Hallu} & \multirow{2}[2]{*}{\textbf{Avg.}} \\ 
    \cmidrule(lr){2-5}\cmidrule(r){6-7}\cmidrule(r){8-9}\cmidrule(r){10-10}
    & \textbf{GQA} & \textbf{MMB} & \textbf{MME$^{C}$} & \textbf{MME$^{P}$} & \textbf{TextVQA} & \textbf{OCRBench} & \textbf{CVBench$^{2D}$} & \textbf{CVBench$^{3D}$} & \textbf{POPE} & \\ 
    \midrule
    \rowcolor{gray!15} Mini-LLaVA & 56.95 & 46.91 & 262 & 1200 & 35.47 & 239 & 39.74 & 55.00 & 85.83 & 48.51 \\

    + \textit{Single} & 59.14$^{\color{red}{1.06}\uparrow}$ & 54.04$^{\color{red}{1.63}\uparrow}$ & 237$^{\color{red}{3}\uparrow}$ & 1154 & 37.84$^{\color{red}{1.81}\uparrow}$ & 265$^{\color{red}{14}\uparrow}$ & 41.36$^{\color{red}{0.21}\uparrow}$ & 57.00$^{\color{red}{4.42}\uparrow}$ & 85.64$^{\color{teal}{0.02}\downarrow}$ & 49.87$^{\color{red}{1.21}\uparrow}$ \\
    + \textit{Double} & 59.19$^{\color{red}{0.60}\uparrow}$ & 53.78$^{\color{red}{4.82}\uparrow}$ & 238$^{\color{red}{9}\uparrow}$ & 1141$^{\color{teal}{66}\downarrow}$ & 38.35$^{\color{red}{2.11}\uparrow}$ & 256$^{\color{red}{1}\uparrow}$ & 42.05$^{\color{red}{0.18}\uparrow}$ & 50.50$^{\color{teal}{6.00}\downarrow}$ & 86.33$^{\color{red}{0.46}\uparrow}$ & 49.18$^{\color{red}{0.64}\uparrow}$ \\
    + \textit{Former} & 58.48$^{\color{red}{4.40}\uparrow}$ & 51.89$^{\color{red}{8.75}\uparrow}$ & 253$^{\color{red}{15}\uparrow}$ & 1180$^{\color{red}{60}\uparrow}$ & 35.67$^{\color{red}{9.16}\uparrow}$ & 261$^{\color{red}{61}\uparrow}$ & 41.22$^{\color{red}{6.28}\uparrow}$ & 50.92$^{\color{teal}{0.41}\downarrow}$ & 85.62$^{\color{red}{1.48}\uparrow}$ & 48.95$^{\color{red}{4.52}\uparrow}$ \\
    + \textit{Latter} & 58.55$^{\color{teal}{0.24}\downarrow}$ & 54.47$^{\color{red}{7.45}\uparrow}$ & 231$^{\color{red}{10}\uparrow}$ & 1144$^{\color{teal}{35}\downarrow}$ & 36.79$^{\color{teal}{0.49}\downarrow}$ & 254$^{\color{red}{13}\uparrow}$ & 38.12$^{\color{teal}{4.52}\downarrow}$ & 51.33$^{\color{teal}{0.59}\downarrow}$ & 86.17$^{\color{red}{0.60}\uparrow}$ & 48.54$^{\color{red}{0.33}\uparrow}$ \\
    + \textit{All} & 59.54$^{\color{red}{1.50}\uparrow}$ & 52.15$^{\color{red}{4.87}\uparrow}$ & 236$^{\color{red}{12}\uparrow}$ & 1200$^{\color{teal}{15}\downarrow}$ & 38.01$^{\color{red}{3.35}\uparrow}$ & 255$^{\color{red}{12}\uparrow}$ & 44.78$^{\color{red}{1.92}\uparrow}$ & 53.08$^{\color{red}{1.62}\uparrow}$ & 86.40$^{\color{red}{1.21}\uparrow}$ & 49.88$^{\color{red}{1.82}\uparrow}$ \\

    \bottomrule
    \end{tabular}
\end{table*}

For external fusion, we provide a combined discussion of modular fusion and direct fusion. As shown in \cref{fig:different_fusion}, there are significant differences in modular fusion between internal and external fusion. In particular, internal fusion requires more parameters due to the inclusion of cross-attention modules within the LLM and projectors for each visual layer. 
Given this distinction in architecture and parameter usage, we simplify our analysis by discussing both modular and direct fusion together for external fusion. The specific parameter sizes for these additional modules in our setup are detailed in the supplementary materials.

For modular fusion in external fusion methods, we employ the recently proposed MMFuser \cite{cao2024mmfuser}, with performance results shown in \cref{tab: external_modular}. Notably, in external fusion, the \textit{Double} and \textit{Triple} are identical due to the derivation of the global visual feature from the 23rd layer in the Mini-LLaVA baseline. For direct fusion in external fusion methods, we adopt a straightforward approach the same as Dense Connector \cite{yao2024dense}. Specifically, for the different layer sets mentioned above, when selecting fewer layers, we fuse information across layers through dimension concatenation. For the \textit{Former}, \textit{Latter}, and \textit{All}, we apply summation followed by averaging to integrate information from different layers. The performance of direct fusion in external fusion method is shown in \cref{tab: ablation on external fusion}. We can conclude the following:

\begin{itemize}

    \item \textbf{Stronger Performance in External Fusion}: Both modular and direct fusion strategies demonstrate superior performance in external fusion compared to internal fusion. Specifically, modular and direct fusion in external fusion achieve performance levels of 49.78 and 49.88 under the \textit{Former} and \textit{All}, respectively. In contrast, the highest internal fusion performance reaches only 48.74. Notably, internal fusion shows a limited advantage over external fusion in the OCR and CV-Centric categories but lags in General and Hallu benchmarks.

    \item \textbf{Direct Fusion Suffices in External Fusion}: In external fusion, direct fusion alone is effective for integrating multi-layer visual features, and adding additional modules does not lead to a notable performance gain. Moreover, adding layers may even lead to performance drops in modular fusion. For instance, \textit{All} performs 0.69 points lower than \textit{Former}. Conversely, the simple averaging approach used in direct fusion achieves optimal results under the \textit{All}, highlighting its effectiveness.
    
    \item \textbf{Higher Performance Variance in Modular Fusion}: Similar to internal fusion, modular fusion in external fusion displays greater performance variance across different layer combinations, even though external fusion keeps the parameter count stable. This finding suggests that modular fusion remains more sensitive to layer selection compared to direct fusion, which shows consistent results regardless of the number of fused layers.
\end{itemize}

\section{Further Analysis}
\label{sec:further_analysis}
To identify an effective approach for multi-layer visual features, we first analyze the generalization of fusion strategies across different configurations. Based on their adaptability, we propose a comprehensive recipe for implementation.  Our experiments vary training data size, visual encoder, and LLM selection to assess the consistency of fusion strategies, verifying whether patterns from \textbf{Section 3} hold across settings.

\begin{figure}[ht]
    \centering
  \includegraphics[width=0.85\columnwidth]{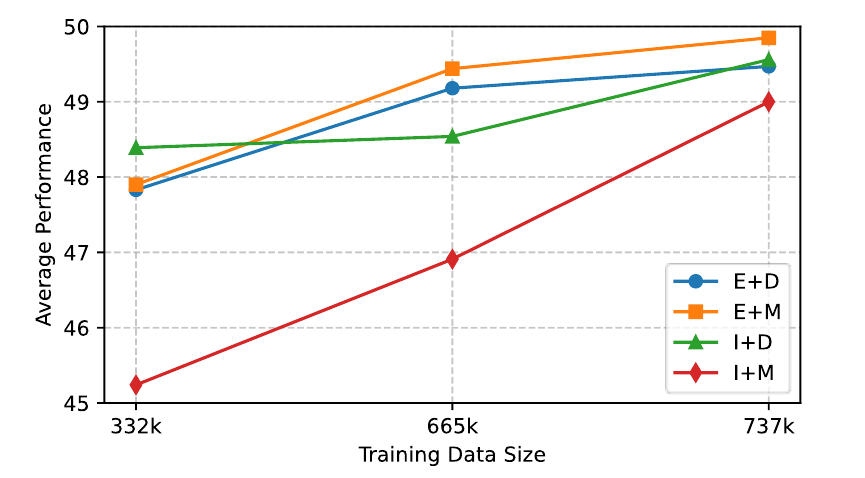}
  \caption{
The performance trend of different fusion strategies adopting the \textit{Triple} visual layers as the training dataset increases.
The abbreviations represent different fusion configurations: \textbf{E} denotes External fusion, \textbf{I} denotes Internal fusion, \textbf{D} stands for Direct fusion, and \textbf{M} indicates Modular fusion.
  }
  \label{fig:different_data_size}
\end{figure}

\subsection{Effect of Data Scale}

Given that different fusion strategies involve varying parameter requirements, exploring the effect of data size on fusion performance is essential. Internal fusion methods \cite{alayrac2022flamingo,chen2024evlm,dai2024nvlm,ye2024mplug} often demand extensive training data.
In contrast, other fusion methods \cite{meng2024deepstack,cao2024mmfuser,yao2024dense} show good results even with smaller datasets. To investigate whether limited training data (558k + 665k) from LLaVA-1.5 constrained performance, we experiment with three different SFT (Supervised Fine-Tuning) training data sizes: 332k, 665k, and 737k. Here, 665k represents the SFT data from LLaVA-1.5, 332k is half of this, and 737k comes from Cambrian-1 \cite{tong2024cambrian}. We conduct detailed ablation experiments across four fusion strategies, with results shown in \cref{fig:different_data_size}. 
It can be observed that while E+D and E+M maintain high performance with data sizes of 665k or more, a notable trend emerges: as data size increases, the performance improvement of I is more pronounced. This suggests that with larger datasets, internal fusion may become a viable option.

\begin{figure}[h]
    \centering
  \includegraphics[width=\columnwidth]{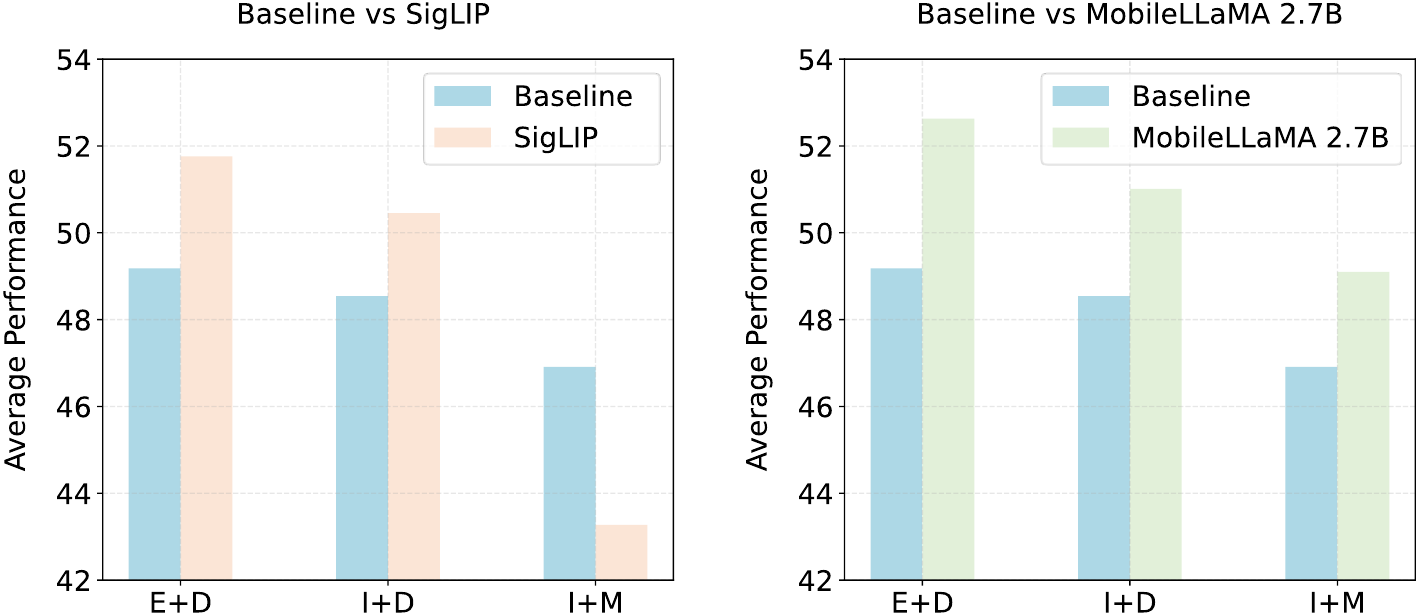}
  \caption{
The performance comparison of different fusion strategies adopting the \textit{Triple} visual layers when replacing model components. The abbreviations represent different fusion configurations: \textbf{E} denotes External fusion, \textbf{I} denotes Internal fusion, \textbf{D} stands for Direct fusion, and \textbf{M} indicates Modular fusion.
  }
  \label{fig:different_component}
\end{figure}

\subsection{Effect of Model Components}
Multimodal large language models offer various options for component selection. For the visual encoder, in addition to the classic CLIP ViT-L, the recently introduced SigLIP has garnered significant attention \cite{zhai2023siglip}. As for the LLM, we chose the 2.7B MobileLLM to explore the performance of larger LLMs. As shown in \cref{fig:different_component}, both external and internal fusion strategies exhibit consistent performance improvements when scaling with more advanced model components, such as the SigLIP visual encoder or the 2.7B MobileLLaMA. Importantly, these results confirm the conclusions drawn from our baseline experiments: external fusion consistently outperforms internal fusion. For instance, in the ``External + Direct Fusion" (E + D) configuration, the average score increases from 49.18 in the baseline to 51.76 with SigLIP and further to 52.63 with MobileLLaMA 2.7B, reinforcing the scalability advantage of external fusion in integrating multi-layer visual features.

Notably, the ``Internal + Modular Fusion" (I + M) configuration underperforms significantly when paired with the SigLIP visual encoder, reaching an average score of only 43.27. This drop suggests that modular methods within internal fusion may struggle with parameter efficiency or may be prone to overfitting, especially when integrating more complex visual encoders.

\section{Conclusion}

This study provides an in-depth analysis of utilizing multi-layer visual features, focusing on two main questions:

For the question 1 in \cref{sec:intro}: \textbf{How can we select visual layers more effectively?}
We found that selecting representative layers from the beginning and middle stages can significantly improve all fusion strategies, especially for detail-sensitive tasks such as OCR and CV-centric tasks. Notably, repeatedly fusing features from the ending stage, such as the 23rd layer, which has already been used as visual tokens, did not provide substantial improvements and may even lead to performance degradation when fused into the model. Similarly, compared to configurations that include early-layer features (e.g., \textit{Former} or \textit{All}), using only \textit{Latter} resulted in weaker performance on specific detail tasks.
In conclusion, the most effective way to select visual layers is to choose one representative visual feature from both the beginning and middle stages, along with the visual tokens generated from the ending stage, forming a comprehensive set of multi-layer visual features.

For the question 2 in \cref{sec:intro}: \textbf{How can we select effective fusion strategies?}
We found that in most cases, external fusion consistently outperforms internal fusion. However, when trained on large datasets, internal fusion shows significant performance improvement, suggesting that it has the potential to approach the effectiveness of external fusion under optimal conditions. Additionally, through experiments with different model configurations and layer selection sets, we observe that direct fusion exhibited greater stability than modular fusion, which introduces more variance.
In summary, the most effective fusion strategy is external direct fusion, as it consistently demonstrates strong performance and excellent generalization across various settings. When a large training dataset is available, internal direct fusion can also be considered as a potential alternative.

\section*{Acknowledgement}
This work was supported in part by NSFC 62302246, ZJNSFC under Grant LQ23F010008. This research was also supported by A*STAR, CISCO Systems (USA) Pte. Ltd and National University of Singapore under its Cisco-NUS Accelerated Digital Economy Corporate Laboratory (Award I21001E0002).

{
    \small
    \bibliographystyle{ieeenat_fullname}
    \bibliography{main}
}
\clearpage
\appendix
\addcontentsline{toc}{section}{Appendix} 
\renewcommand \thepart{} 
\renewcommand \partname{}
\part{\Large{\centerline{Appendix}}}

\section{Visualization of Cross-Attention Modules}
\begin{figure}[h]
    \centering
  \includegraphics[width=\columnwidth]{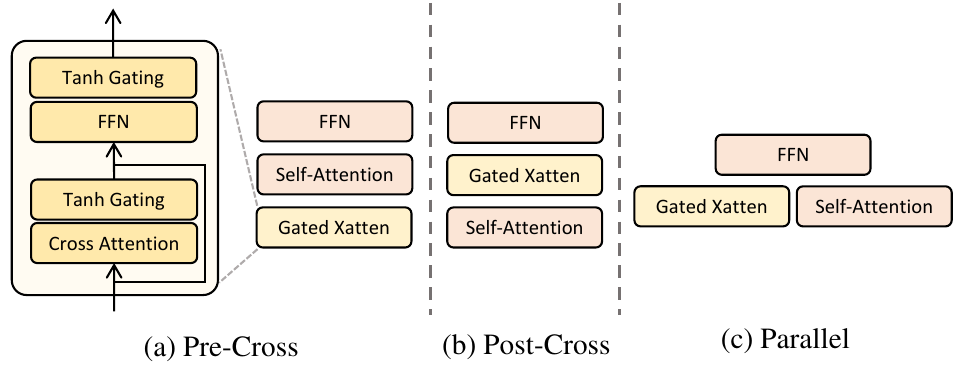}
  \caption{
Architecture Comparisons between three Current Internal Modular Fusion Strategies. 
  }
  \label{fig:internal_moduler_fusion}
\end{figure}

\cref{fig:internal_moduler_fusion} illustrates the architectures of different Internal Modular Fusion used in our experiments. The "Gated Xatten" module, adapted from \cite{alayrac2022flamingo}, is introduced as a new component in these architectures. Based on where the "Gated Xatten" module is inserted, we identify three distinct architectures: pre-cross, post-cross, and parallel. These architectures are further compared in terms of their efficiency in fusing multi-layer visual features, as detailed in Tab. \ref{tab: ablation on modular fusion strategies}.

\vspace{-1mm}

\section{Training Datasets and Evaluation Benchmarks}

\subsection{Training Datasets}

For the training data, we utilize three main datasets:
\begin{enumerate}
    \item The first dataset is used for pretraining. This dataset comprises a subset of 558K LAION-CC-SBU \cite{laion,cc,sbu} image-text pairs with BLIP-generated captions \cite{li2022blip}, which is the same as the first stage of the LLaVA-1.5 \cite{liu2024improved} pre-training.
    \item The second and third datasets are used for instruction tuning. The second dataset, which is the primary fine-tuning dataset used in most of our experiments, consists of a 665K instruction-following data summarized by LLaVA-1.5. The third part, derived from Cambrian-1 \cite{tong2024cambrian}, builds upon the 665K instruction-following dataset of LLaVA-1.5 by adding a small number of OCR and chart data. The detailed composition of these datasets can be found in \cref{tab:detail_ft}.
\end{enumerate}

\begin{table}[t!]
\caption{
The mixture detail of fine-tuning dataset for LLaVA-1.5 665K and Cambrian-1 737K.
}
\label{tab:detail_ft}
\centering \footnotesize
  \renewcommand\tabcolsep{15.6pt}
\begin{tabular}{l l}
\toprule
Data & Size \\
\midrule
LLaVA~\cite{liu2024visual} & 158K \\
+ ShareGPT~\cite{sharegpt} & 40K \\
+ VQAv2~\cite{goyal2017vqav2} & 83K \\
+ GQA~\cite{hudson2019gqa} & 72K \\
+ OKVQA~\cite{okvqa} & 9K \\
+ OCRVQA~\cite{mishra2019ocrvqa} & 80K \\
+ A-OKVQA~\cite{schwenk2022okvqa} & 66K  \\
+ TextCaps~\cite{sidorov2020textcaps} & 22K  \\
+ RefCOCO \cite{kazemzadeh2014referitgame,mao2016generation} & 48K \\
+ VG~\cite{krishna2017visual} & 86K  \\
LLaVA-1.5 Total & 665K  \\
\midrule
+ AI2D~\cite{kembhavi2016diagram} & 16K \\
+ DocVQA~\cite{mathew2021docvqa} & 15K \\
+ DVQA~\cite{kafle2018dvqa} & 13K \\
\midrule
Cambrian-1 Total & 737K  \\
\bottomrule
\end{tabular}
\end{table}

\begin{table*}[!htpb]
    \centering
    \caption{Comparison on Different Training Datasets. \textbf{Note:} E, I, D, and M represent External Fusion, Internal Fusion, Direct Fusion, and Modular Fusion, respectively.}

    \scalebox{0.85}{
    \renewcommand\tabcolsep{4pt}
    \begin{tabular}{lc cccc cc cc c c}
    \toprule
    \multirow{2}{*}{\textbf{FT}} & \multirow{2}{*}{\textbf{PT+IT}} & \multicolumn{4}{c}{\textbf{General}} & \multicolumn{2}{c}{\textbf{OCR}} & \multicolumn{2}{c}{\textbf{CV-Centric}} & \textbf{Hallu} & \multirow{2}{*}{\textbf{Avg.}} \\ 
    \cmidrule(lr){3-6} \cmidrule(lr){7-8} \cmidrule(lr){9-10} \cmidrule(lr){11-11}

    & & \textbf{GQA} & \textbf{MMB} & \textbf{MME$^{C}$} & \textbf{MME$^{P}$} & \textbf{TextVQA} & \textbf{OCRBench} & \textbf{CVBench$^{2D}$} & \textbf{CVBench$^{3D}$} & \textbf{POPE} & \\ 
    \midrule

    \multirow{3}{*}{\textbf{E + D}} 
    & 558k + 332k     & 56.04 & 49.14 & 224 & 1126 & 34.56 & 266 & 43.96 & 50.00 & 85.91 & 47.83 \\
    & 558k + 665k    & 59.19 & 53.78 & 238 & 1141 & 38.35 & 256 & 42.05 & 50.50 & 86.33 & 49.18 \\
    & 558k + 737k     & 59.73 & 52.84 & 208 & 1202 & 39.21 & 285 & 41.53 & 50.58 & 86.71 & 49.47 \\
    
    \midrule
    \multirow{3}{*}{\textbf{E + M}} 
    & 558k + 332k     & 54.90 & 50.77 & 243 & 1055 & 34.16 & 250 & 45.53 & 51.58 & 86.01 & 47.90 \\
    & 558k + 665k    & 58.43 & 52.66 & 225 & 1173 & 36.25 & 262 & 45.78 & 52.83 & 86.03 & 49.44 \\
    & 558k + 737k     & 58.82 & 51.11 & 241 & 1211 & 37.19 & 280 & 44.11 & 51.92 & 86.78 & 49.85 \\
    
    \midrule
    \multirow{3}{*}{\textbf{I + D}} 
    & 558k + 332k     & 55.06 & 52.14 & 238 & 1046 & 33.74 & 219 & 48.77 & 55.83 & 86.04 & 48.39 \\
    & 558k + 665k    & 58.59 & 47.47 & 223 & 1207 & 36.24 & 255 & 41.87 & 53.08 & 85.87 & 48.54 \\
    & 558k + 737k     & 58.09 & 49.66 & 244 & 1188 & 37.05 & 272 & 43.52 & 50.50 & 86.13 & 49.56 \\
    
    \midrule
    \multirow{3}{*}{\textbf{I + M}} 
    & 558k + 332k     & 52.26 & 43.56 & 234 & 1027 & 31.12 & 236 & 43.01 & 48.08 & 84.96 & 45.24 \\
    & 558k + 665k    & 57.56 & 49.66 & 212 & 1163 & 34.06 & 255 & 38.66 & 47.42 & 84.69 & 46.91 \\
    & 558k + 737k     & 58.09 & 51.11 & 241 & 1172 & 35.08 & 272 & 46.96 & 47.75 & 86.09 & 49.00 \\

    \bottomrule
    \end{tabular}}
    \label{tab: ablation on fusion strategies}
\end{table*}

\begin{table*}[!htpb]
    \centering
    \caption{Comparison on Different MLLM Components.}
    \scalebox{0.85}{
    \renewcommand\tabcolsep{4pt}
    \begin{tabular}{ll cccc cc cc c c}
    \toprule
    \multirow{2}{*}{\textbf{FT}} & \multirow{2}{*}{\textbf{Component}} & \multicolumn{4}{c}{\textbf{General}} & \multicolumn{2}{c}{\textbf{OCR}} & \multicolumn{2}{c}{\textbf{CV-Centric}} & \textbf{Hallu} & \multirow{2}{*}{\textbf{Avg.}} \\ 
    \cmidrule(lr){3-6} \cmidrule(lr){7-8} \cmidrule(lr){9-10} \cmidrule(lr){11-11}
    & & \textbf{GQA} & \textbf{MMB} & \textbf{MME$^{C}$} & \textbf{MME$^{P}$} & \textbf{TextVQA} & \textbf{OCRBench} & \textbf{CVBench$^{2D}$} & \textbf{CVBench$^{3D}$} & \textbf{POPE} & \\ 
    \midrule

    \multirow{3}{*}{\textbf{E + D}} 
    & Baseline    & 59.19 & 53.78 & 238 & 1141 & 38.35 & 256 & 42.05 & 50.50 & 86.33 & 49.18 \\
    & +SigLIP     & 60.69 & 53.26 & 219 & 1245 & 45.98 & 302 & 41.31 & 57.92 & 86.84 & 51.76 \\
    & +MobileLLaMA 2.7B     & 61.39 & 59.28 & 238 & 1293 & 42.54 & 280 & 45.50 & 55.17 & 87.41 & 52.63 \\ 
    \midrule
    \multirow{3}{*}{\textbf{I + D}} 
    & Baseline     & 58.59 & 47.47 & 223 & 1207 & 36.24 & 255 & 41.87 & 53.08 & 85.87 & 48.54 \\
    & +SigLIP    & 59.32 & 54.73 & 230 & 1162 & 41.66 & 272 & 42.91 & 55.33 & 86.02 & 50.45 \\
    & +MobileLLaMA 2.7B     & 60.56 & 58.33 & 235 & 1283 & 39.94 & 266 & 39.06 & 54.42 & 86.62 & 51.01 \\     
    \midrule
    \multirow{3}{*}{\textbf{I + M}} 
    & Baseline     & 57.56 & 49.66 & 212 & 1163 & 34.06 & 255 & 38.66 & 47.42 & 84.69 & 46.91 \\
    & +SigLIP     & 50.82 & 45.45 & 262 & 1029 & 16.43 & 134 & 44.94 & 52.67 & 81.52 & 43.27 \\
    & +MobileLLaMA 2.7B     & 58.10 & 50.77 & 246 & 1233 & 33.09 & 246 & 46.72 & 50.00 & 86.22 & 49.10 \\     
    \bottomrule
    \end{tabular}}
    \label{tab: ablation on components}
\end{table*}

\subsection{Evaluation Benchmarks}

To conduct a comprehensive evaluation of MLLMs across different configurations, we have prepared seven distinct benchmarks, categorized into four types: \textit{General}, \textit{OCR}, \textit{CV-Centric}, and \textit{Hallucination}. The capabilities evaluated by each category are as follows:

\begin{itemize}
    \item \textbf{General}: Evaluates the general capabilities of multimodal models, including cognition and perception. Benchmarks in this category include:
    GQA~\cite{hudson2019gqa}, MMBench (MMB)~\cite{liu2025mmbench}, and MME~\cite{fu2023mme}, which is further divided into MME Cognition (MME$^C$) and MME Perception (MME$^P$).
    
    \item \textbf{OCR}: Evaluates the model's performance in text recognition and understanding tasks. Benchmarks in this category include:
    TextVQA~\cite{singh2019textvqa} and OCRBench~\cite{liu2023ocrbench}.
    
    \item \textbf{CV-Centric}: This category focuses on better evaluating visual representations in an integrated multimodal setting. Benchmarks in this category include:
    CV-Bench~\cite{tong2024cambrian}, which is further divided into CV-Bench$^{2D}$ and CV-Bench$^{3D}$.
    
    \item \textbf{Hallucination}: Evaluates the model's ability to generate accurate and truthful information, avoiding hallucinations. The benchmark in this category is:
    POPE~\cite{li2023pope}.
\end{itemize}

\section{More Detail about Results}
In Section 5, we present the performance differences of various fusion strategies under the \textit{Triple} layer selection set when dealing with different data scales and model components, as shown in Fig. \ref{fig:different_data_size} and Fig. \ref{fig:different_component}. To provide a more comprehensive understanding of the performance differences, we include the complete evaluation results in \cref{tab: ablation on fusion strategies} and \cref{tab: ablation on components}. 

\begin{figure}[htpb]
    \centering
  \includegraphics[width=0.9\columnwidth]{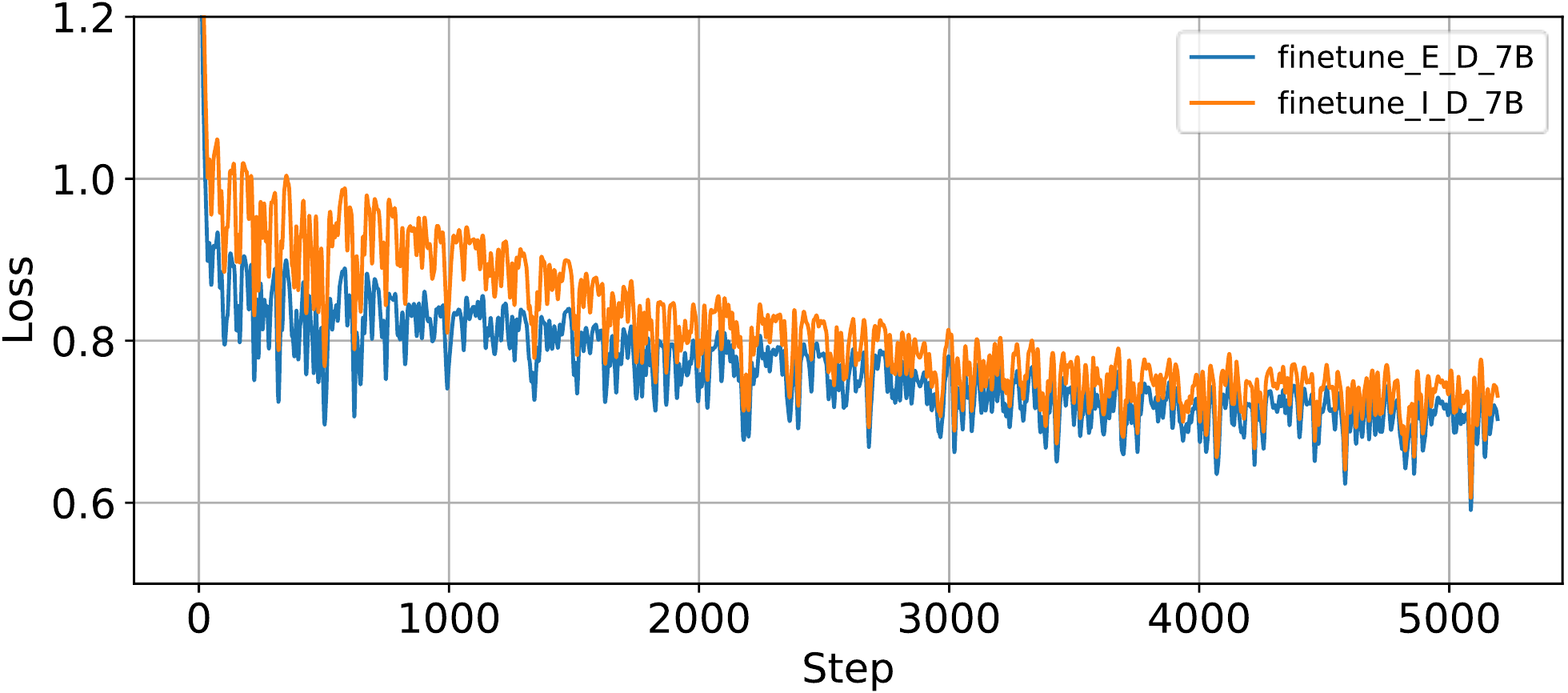}
  \caption{
Loss curves for different fusion strategies.
  }
  \label{fig:layer_selection_sup}
\end{figure}

We further explored different layer combinations and experimented with a larger language model (Vicuna 1.5 7B \cite{chiang2023vicuna}). As shown in \cref{tab:more}, fusing visual layer \{3\} alone yields lower performance than fusing visual layer \{18\}, indicating that the earlier layer (layer 3) has a lesser impact. Comparing the two fusion methods, external direct fusion shows greater performance gains (average score: 49.18 \(\to\) 52.63 \(\to\) 58.62) than internal direct fusion (average score: 48.54 \(\to\) 51.01 \(\to\) 54.37). Coupled with the weaker results on smaller datasets (see \cref{fig:different_data_size}), this phenomenon suggests that internal fusion may disrupt intrinsic feature distributions within LLM. However, by more closely integrating ViT’s object-focused features with the LLM’s higher-level abstractions, internal fusion holds substantial theoretical promise. As illustrated in \cref{fig:layer_selection_sup}, the training loss for internal fusion gradually converges but continues to improve, implying that, with sufficient data, this approach has the possibility of outperforming simpler external fusion strategies.
\begin{table}[hb!]
\centering
\caption{Comparison of performance across different configurations.}
\vspace{-3pt}
\begin{adjustbox}{width=0.47\textwidth}
\begin{tabular}{lccccc}
\hline
\textbf{Model} & \textbf{Avg.} & General & OCR & CV-Centric & Hallu\\
\hline
\rowcolor{gray!15} Mini-LLaVA
& 48.51 & 49.15 & 29.69 & 47.37 & 85.83\\

E+D \{3\}
& 48.26 & 49.24 & 29.53 & 46.17 & 86.01\\

E+D \{3,18,23\} Vicuna 1.5 7B
& 58.62 & 59.57 & 39.84 & 61.39 & 86.84\\

I+D \{3,18,23\} Vicuna 1.5 7B
& 54.37 & 57.85 & 30.93 & 55.84 & 84.40 \\
\hline
\end{tabular}
\end{adjustbox}
\label{tab:more}
\end{table}

\section{Limitations}
In this study, we examine model scaling by parameter and dataset size. Specifically, for LLMs, our experiments are limited to a maximum scale of 7B parameters. Similarly, for datasets, we constrain the scale to about 1M samples. While these limits are smaller compared to the largest LLMs and datasets available in the field, our exploration still holds significant value due to the practical relevance and commonality of such scales in many works \cite{cao2024mmfuser,yao2024dense,lu2024deepseek}.

\end{document}